\documentclass[final,5p,times,twocolumn]{elsarticle}

\usepackage{amssymb}

\usepackage{caption}
\usepackage{hyperref} 
\hypersetup{
colorlinks=true,
linkcolor=black,
citecolor=black
}
\usepackage{mathrsfs}
\usepackage{amsmath}
\usepackage{amsfonts,amssymb} 
\usepackage{verbatim}
\usepackage{tikz}
\usepackage{placeins}
\usepackage{booktabs}  
\usepackage{textcomp}
\usetikzlibrary{arrows,backgrounds}
\usepackage{tabularx}
\usepackage[T1]{fontenc}
\usepackage{balance}

\usepackage{fancyhdr}

\begin{document}

\begin{frontmatter}

\title{Probabilistic Decomposition Transformer for Time Series Forecasting}

\author{Junlong Tong}
\ead{jltong@seu.edu.cn}
\author{Liping Xie\corref{cor1}}
\ead{lpxie@seu.edu.cn}
\author{Wankou Yang}
\ead{wkyang@seu.edu.cn}
\author{Kanjian Zhang}
\ead{kjzhang@seu.edu.cn}
\cortext[cor1]{Corresponding author}

\address{}
\address{}
\address{School of Automation, Southeast University}

\begin{abstract}
    Time series forecasting is crucial for many fields, such as disaster warning, weather prediction, and energy consumption.
    The Transformer-based models are considered to have revolutionized the field of sequence modeling.
    However, the complex temporal patterns of the time series hinder the model from mining reliable temporal dependencies.
    Furthermore, the autoregressive form of the Transformer introduces cumulative errors in the inference step.
    In this paper, we propose the probabilistic decomposition Transformer model that combines the Transformer with a conditional generative model,
    which provides hierarchical and interpretable probabilistic forecasts for intricate time series.
    The Transformer is employed to learn temporal patterns and implement primary probabilistic forecasts, 
    while the conditional generative model is used to achieve non-autoregressive hierarchical probabilistic forecasts by introducing latent space feature representations.
    In addition, the conditional generative model reconstructs typical features of the series, such as seasonality and trend terms, from probability distributions in the latent space to enable complex pattern separation and provide interpretable forecasts.
    Extensive experiments on several datasets demonstrate the effectiveness and robustness of the proposed model, indicating that it compares favorably with the state of the art.

\end{abstract}

    
\end{frontmatter}
\thispagestyle{fancy} 
\fancyhead{} 
\renewcommand\headrulewidth{0pt}
\fancyfoot{}
\renewcommand{\thefootnote}{\fnsymbol{footnote}}

\section{Introduction}

In the era of big data, the explosive growth of data has forced the development of data mining.
Time series information, as one of the most common data, has captured widespread attention for the mining of its intrinsic patterns.
Time series data mining refers to discovering potential features to provide decision making, such as prediction, classification, and anomaly detection, by learning historical data in chronological order.
Among them, the importance and application of time series forecasting have been demonstrated in modern society such as commodity demand forecasting \cite{zi2021tagcn}, energy consumption \cite{tong2022hourly}, traffic planning \cite{ali2021exploiting}, and financial analysis \cite{sun2021multi}.

% Time series forecasting has demonstrated its importance and applications in modern society such as commodity demand forecasting, weather forecasting, traffic planning, and disaster warning.
% In these real-world applications,  
% Probabilistic forecasts take the form of a predictive probability distribution over future
Classical statistical methods have been well used for time series forecasting, such as ARIMA \cite{box1970time} models and state space models (SSMs) \cite{hyndman2002state}.
These methods usually incorporate prior knowledge of the time series, such as trend and seasonality characteristics, and use them for decision making and interpretation.
% However, they are difficult to be useful in forecasting complex time series data due to their inability to exploit time-dependent features.
However, statistical-based methods fail to capture the relationship between the covariates and the target series,  limiting the effectiveness of predicting intricate time series.

In recent years, deep prediction models have been significantly developed to tackle large-scale complex time series forecasting problems.
For example, models based on recurrent neural networks (RNNs) \cite{Chung2015latent,fraccaro2016sequential} and self-attention mechanisms \cite{vaswani2017attention,li2019enhancing} are widely used for time series forecasting.
Models based on the self-attention mechanism simultaneously calculate the attention scores between any two time points, assigning different weights according to the importance of each part of the input data. 
Compared with the circular feature of RNN, the self-attention mechanism-based model processes the historical time series data at once and realizes the information interaction at different moments, which can maintain the long-term dependencies.
However, forecasting models based on RNN and self-attention mechanism follow an autoregressive form and usually employ the teacher-forcing strategy \cite{williams1989learning,lamb2016professor}, i.e., providing ground truth at each moment in the training phase, to improve convergence and generalization.
However, the strategy causes inconsistencies between the training and inference phases, leading to exposure bias \cite{bengio2015scheduled} in the inference phase.

In addition, the complex temporal patterns of the time series prevent the models from mining reliable temporal dependencies. 
To separate the complex patterns, the concept of series decomposition \cite{hyndman2018forecasting,cleveland1990stl} is introduced into time series analysis, which assumes that the sequence consists of components based on prior knowledge and seeks to separate them.
However, due to the specificity of the forecasting task, only the input historical series can be pre-processed, ignoring the interaction with the forecast information and lacking flexibility.

In this work, we propose to combine the strengths of the Transformer architecture and conditional generative model, for hierarchical and interpretable probabilistic forecasting.
% The model uses Transformer for temporal feature extraction and primary forecasting, where the process is autoregressive in forecasting the probability distribution parameters of the time series.
The Transformer \cite{vaswani2017attention} is used for temporal feature extraction and primary probabilistic forecasting, where the probability distribution parameters of the time series are forecast by an autoregressive process.
In addition, the probability distribution parameters are used as conditional information for probabilistic encoding and reconstruction of the prediction using a variational inference \cite{kingma2013auto,rezende2014stochastic} generative model.
The hierarchical probabilistic forecasting method can effectively mitigate the exposure bias brought by the autoregressive Transformer model by introducing a conditional generative model to constrain the primary forecasting results at the sequence level.
In addition, the probabilistic decoder is designed based on the prior knowledge of the time series to reconstruct the subsequence of different features from the latent space to achieve the separation of intricate patterns and interpretable forecasts, where the probabilistic decoder combines historical sequences and primary forecasting result for information interaction.

The principal contributions of this work can be summarized as follows:
\begin{itemize}
\item 
% To tackle the complex temporal patterns of time series, we design a jointly learned series decomposition framework and adversarial Transformer framework for time series prediction.
An efficient time series forecasting model called Probabilistic Decomposition Transformer is proposed in this work, where the model combines the Transformer architecture and generative model based on variational inference for hierarchical probabilistic forecasting.
% We propose an effective time series forecasting model called Probabilistic Decomposition Transformer, which the model combines the Transformer architecture and generative model based on variational inference for hierarchical probabilistic forecasting.
\item 
% We design a novel probabilistic decomposition framework for interpretable forecasts, which utilizes a variational autoencoder to decompose the distribution of the input sequence into several Gaussian distributions in the latent space, and reconstructs several subsequences of known distributions by sampling the Gaussian distributions.
We design a novel probabilistic decomposition framework, where the probabilistic decoder reconstructs typical features of sequences from probability distributions in the latent space to achieve separation of intricate temporal patterns and provide interpretable forecasts.

\item 
The experiments demonstrate that the proposed model is effective in reducing the exposure bias of autoregressive forecasting, showing that it compares favorably with the state-of-the-art models.

\end{itemize}

\section{Related Work}
\subsection{Time Series Forecasting}
The early methods used for time series forecasting are mainly statistical models, such as ARIMA \cite{box1970time} and exponential smoothing \cite{billah2006exponential}. 
However, statistical models have difficulties capturing complex temporal patterns,  which limits their application to complex scenarios.
Recently, deep neural networks have been applied to time series forecasting thanks to the powerful feature representation capabilities and scalable structure.
The most prominent of the deep models is the recurrent neural networks (RNNs) \cite{wen2017multi,hewamalage2021recurrent}.
% The most notable of which is recurrent neural networks (RNNs) \cite{wen2017multi,hewamalage2021recurrent}.
DeepAR \cite{salinas2020deepar} presented an autoregressive RNNs method for probabilistic distribution modeling. 
DSSM \cite{rangapuram2018deep} combined RNNs with a state space model for probabilistic time series forecasting.
Besides, convolutional neural networks (CNNs) also have been proposed for time series forecasting.
LSTNet \cite{lai2018modeling} combined CNNs with skip-RNN to extract local patterns among variables and capture long-term dependency based on time series trends.
Bai et al. \cite{bai2018empirical} proposed the temporal convolution network (TCN),  a sequence model based on causal convolutions and dilated convolutions, which has been used for time series forecasting \cite{sen2019think,chen2020probabilistic}.
% Recently, Transformer-based model has been proposed to solving time series forecasting. 

In recent years, Transformer-based 
 \cite{vaswani2017attention,wen2022transformers} models have attracted the attention of researchers. 
Most of the work is dedicated to modifying the self-attention mechanism \cite{li2019enhancing,xu2021autoformer,zhou2022fedformer} or to improving the Transformer architecture \cite{liu2021pyraformer,zhou2021informer}.
For example, Li et al. \cite{li2019enhancing} first applied Transformer to time series forecasting, where a convolution Transformer is proposed to improve local processing power and a log-sparse strategy is designed for breaking the memory bottleneck.
SSDNet \cite{lin2021ssdnet} conducted the Transformer to learn the parameters of the state space model to provide probabilistic and interpretable forecasts.
% In this work, we utilize Transformer for autoregressive probability forecasting.
In this work, Transformer is employed to extract temporal patterns and implement primary probabilistic forecasting.

\subsection{Decomposition of Time Series}
% As time series contain multiple underlying patterns typically, one of effective way is to decompose them into several components, each of which corresponds to a particular pattern.
Time series typically contain multiple underlying patterns that increase the complexity of prediction. 
Time series decomposition \cite{west1997time,cleveland1990stl} is regarded as an effective method for pattern separation, which is to decompose a series into several components, each of which corresponds to a particular pattern.
For time series forecasting tasks, a variety of decomposition strategies have been proposed for mining historical volatility over time.
For instance, Prophet \cite{taylor2018forecasting} provides a trend-seasonality decomposition, and DeepGLO\cite{sen2019think} implements a global matrix factorization model for high-dimensional time series.

In recent years, learning-based decomposition strategies have attracted the attention of researchers.
For instance, Oreshkin et al. \cite{oreshkin2019n} proposed the N-Beats model with basis expansion, an interpretable prediction framework with fully-connected layer structures. 
Nguyen et al. \cite{nguyen2021temporal} proposed a temporal latent autoencoder for the nonlinear decomposition of multivariate time series.
Furthermore, in Autofomer\cite{xu2021autoformer} and FEDformer\cite{zhou2022fedformer}, decomposition based on average pooling operation is conducted to highlight the trend term, where the decomposition module is considered as the inner block of forecasting model to decompose series progressively.
This paper proposes a new decomposition idea, which obtains primary prediction results through Transformer model, then uses conditional generation model for probabilistic decomposition to reconstruct subsequences of different features from potential space. The process effectively realizes the interaction between historical data and prediction data, breaks the information barrier, and improves the reliability of decomposition results.

\subsection{Generative Model for Time Series}
Generative models have been extensively used for time series data mining tasks, such as time series imputation \cite{zhang2021missing,miao2021generative}, time series data generation \cite{yoon2019time,jeha2021psa,han2020sample}.
For forecasting tasks, there have been many generative models based on variational inference \cite{kingma2013auto,sohn2015learning} introduced to conditional probability modeling.
Fortuin et al. \cite{fortuin2020gp} proposed a VAE-based method for deep probability time series imputation.
Tang et al. \cite{tang2021probabilistic} proposed to combine the SSM model with the attention mechanism to achieve non-autoregressive probability forecasting by conditional variational inference.
Li et al. \cite{li2021causal} proposed a reformulated VAE framework for time series disease forecasts.
Besides, a recent approach based on diffusion model \cite{rasul2021autoregressive} is proposed to autoregressive time series forecasting.
In this work, a conditional generative model is employed to tackle the conditional probability of the primary forecasting result.

\section{Preliminaries}
\subsection{Problem Definition}
% Let $\{Y_{i,1:t_0}\}_{i=1}^S$ represent a set of \textit{S} related univariate time series, 
Let $\{Y_{i,1:t_0}\}_{i=1}^N$ represent a set of $N$ related univariate time series, 
where $Y_{i,1:t_0}=[y_{i,1},y_{i,2},...,y_{i,t_0}]\in\mathbb{R}^{1\times t_0}$ and $y_{i,t}$ is the value of $i$th time series at time $t$.
In addition, assume that $X_{i,1:t_0}\in\mathbb{R}^{k\times t_0}$ denotes the $k$-dimensional covariate series corresponding to $Y_{i,1:t_0}$, where the covariates can be static (i.e., time-independent features such as serial number) or dynamic (i.e., time-related features such as time period of the day).

Specifically, the task of this work is to model the conditional distribution $p(Y_{i,t_0+1:t_0+\tau}|Y_{i,1:t_0},X_{i,1:t_0+\tau};\Phi)$, where the future covariates are known.
The conditional distribution can be reduced by a step-by-step forecasting task, where the problem can be denoted as follows:\\
\begin{equation}
\begin{split}\label{Eq:model}
    &p(Y_{i,t_0+1:t_0+\tau}|Y_{i,1:t_0},X_{i,1:t_0+\tau};\Phi)\\
    &~~~~~= \prod_{t=t_0+1}^{t_0+\tau}p(Y_{i,t}|Y_{i,1:t-1},X_{i,1:t};\Phi),
\end{split}
\end{equation}
where $\Phi$ is parameters of the model trained on all $N$ related univariate time series, and the input of the proposed model at time $t$ is a concatenated vector of $Y_{t-1}$ and $X_t$.
Besides, the time range $[1,t_0]$ and $[t_0+1,t_0+\tau]$ are referred as conditional range and prediction range respectively, to be consistent with the literature \cite{li2019enhancing,salinas2020deepar}, where $t_0$ represents the forecasting start moment and $\tau$ represents the forecasting horizon. 
% The subscript $i$ will be omitted in the rest of the paper for simplicity.
Moreover, owing to all univariate time series being trained with the same model, the identification $i$ to distinguish different time series will be omitted in the latter.

\subsection{The Transformer}
% Structurally, Transformer-based models are encoder-decoder architectures based mainly on the self-attention mechanism, where the self-attention mechanism enables the Transformer to capture long-term dependencies and focus on important patterns of the context.
Structurally, Transformer-based models \cite{vaswani2017attention,devlin2018bert} are encoder-decoder architectures based mainly on the self-attention mechanism, which enables the models to capture long-term dependencies and focus on important patterns.
% In practice, multi-head self-attention is usually adopted to the Transformer to improve the model fitting performance.
In practice, multi-headed self-attribution is usually applied to the Transformer to improve the model fitting performance.
For the self-attention model with $H$ head, the outputs $O_1,...,O_H$ are concatenated and linearly projected to $O$, i.e., 
$O=concat(O_1,...,O_H)W^O$, where $W^O$ is the parameter of the linear projection.
% The output of multi-head self-attention can be shown in Equation \ref{Att}.
For $h\in[1,H]$, multi-head self-attention output $O_h$ can be expressed as follows:
\begin{equation}\label{Att}
\begin{split}
O_h =& Attention(Q_h , K_h, V_h) \\
=& softmax(\frac{Q_hK_h^T}{\sqrt{d_k}})V_h,
\end{split}
\end{equation}
where $Q_h=QW_h^Q$, $K_h=KW_h^K$, and $V_h=VW_h^V$ are the projections of query, key, and value with learnable parameters $W_h^Q$, $W_h^K$, and $W_h^V$, respectively.

Compared with the recurrent neural network \cite{hochreiter1997long,cho2014learning} that passes memory states step by step, 
the self-attention mechanism calculates the attention scores of any two moments simultaneously, 
which improves the learning ability of long-term dependencies as well as optimizes parallel performance.
Besides, a masking mechanism is introduced to the decoder to obscure future information for preventing information leakage.
% Then 

\subsection{Variational Autoencoder}
% Variational inference is a parameter estimation method tackling the conditional probability with hidden variables, which approximates the conditional probability by introducing a variational distribution and usually implemented by the variational autoencoder (VAE).
% Variational inference is a parameter estimation method for dealing with the  conditional probability containing hidden variables by introducing a variational distribution, usually implemented by the variational autoencoder (VAE).
Variational inference is a parameter estimation method for dealing with the conditional probability containing hidden variables by introducing a variational distribution, usually implemented by the variational autoencoder (VAE).
The VAE \cite{kingma2013auto} is a deep generative model designed to learn the probability distribution of existing datasets and generate new data through sampling from the learned data distribution. 
Since the form of the probability distribution of existing data is unknown, VAE encodes the input data into the latent space, constructs the probability distribution of the latent variables, samples from the latent variables and reconstructs new data. 

\section{Probabilistic Decomposition Transformer}

\subsection{Model Architecture}
In brief, the aim of the Probabilistic Decomposition Transformer (PDTrans) is to build a hierarchical probabilistic forecasting model 
to provide accurate probabilistic and interpretable forecasts.

\begin{figure*}[htb]
    \centering
    \includegraphics[width=0.98\textwidth]{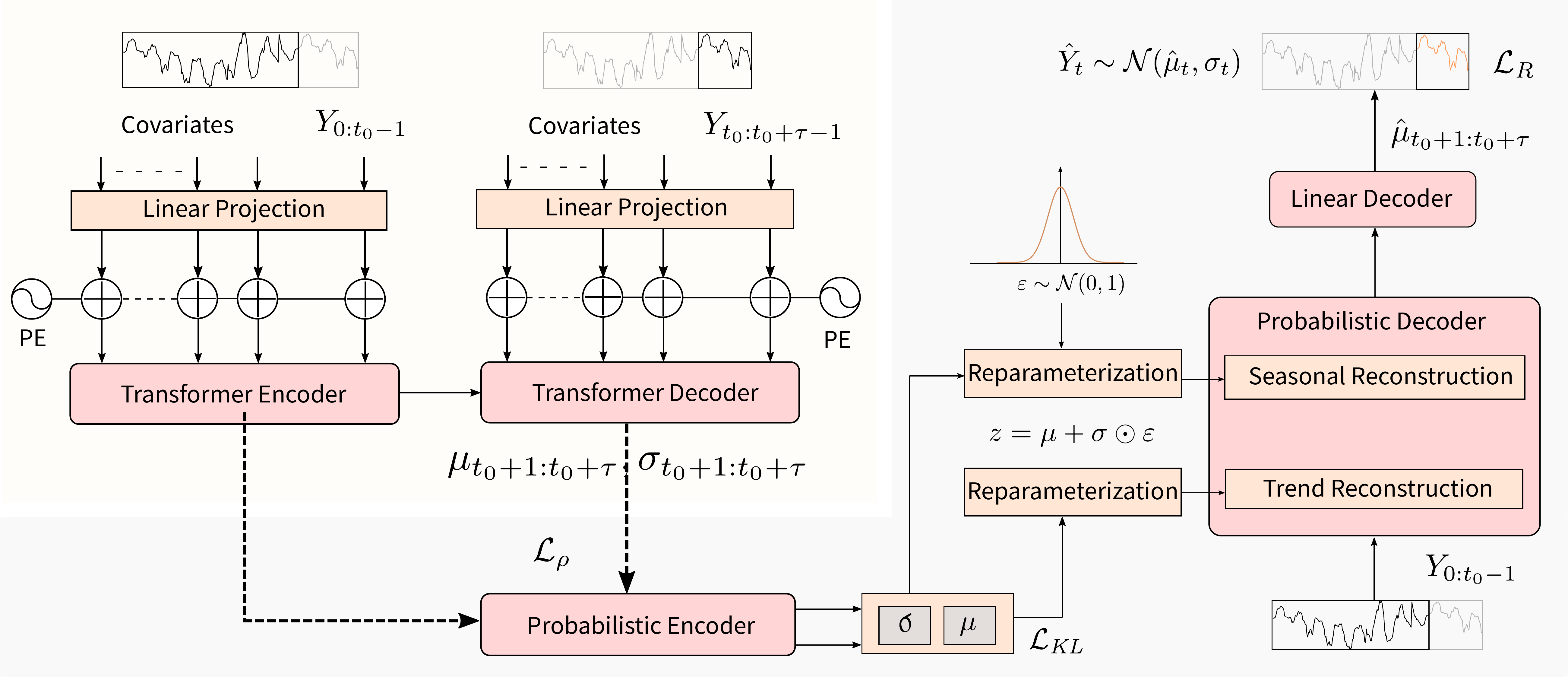}\\
    \caption{Architecture of the PDTrans model. (left) Transformer model for primary autoregressive probabilistic forecasts. (right) Conditional generative model for sequence-level probabilistic forecasts and pattern separation.}
    \label{Fig:PDT}
\end{figure*}

The architecture of the proposed model is illustrated in Figure \ref{Fig:PDT}, which contains two important components: Transformer and conditional generative model.
% The Transformer is employed for primary autoregressive probabilistic forecasts and conditional generative model
The Transformer learns temporal patterns and implements autoregressive probabilistic forecasting, a process that outputs the parameters of the probability distribution to provide primary forecasting.
The function of the conditional generative model is twofold.
The first function is to model the primary forecasting distribution through variational inference to achieve hierarchical forecasting, 
which can mitigate the impact of exposure bias in the autoregressive prediction process.
Reconstructing typical features of sequence from the probability distribution of the latent space is another function,
where a probabilistic decoder is employed to achieve the separation of complex patterns and provide interpretable forecasts.

\subsection{Autoregressive Probabilistic Forecasts}
We attempt to model the conditional probabilities shown in Equation \ref{Eq:model} and assume that 
each conditional probability obeys a specific probability distribution that can be represented 
by a learnable likelihood function, as shown in the following: 
\begin{equation}
\begin{split}\label{Eq:model2}
    p(Y_{t}|Y_{1:t-1},X_{1:t};\Phi)=l(Y_{t}|\Phi_t),
\end{split}
\end{equation}
where $l(Y_{t}|\Phi_t)$ is the likelihood and $\Phi_t$ is the parameter of the likelihood at moment $t$.
The likelihood should be selected based on the statistical properties of the data to accurately approximate the true distribution.
For real-world datasets, the Gaussian likelihood is usually chosen to approximate the conditional probability distribution.

% The Transformer is employed to tackle the problem, where the 
Since the Transformer has made a splash in the field of sequence modeling, this work employs Transformer to tackle the autoregressive probabilistic forecasting problem.
The encoder of Transformer takes the linear projection of history sequence and covariates as input, where the linear projection is applied to learnable embedding, as shown in Figure \ref{Fig:PDT} (left).
Besides, the position information is represented by adding the position encode (PE) to the output of the embedding layer.
The decoder of Transformer outputs the parameters of the likelihood function at each moment in an autoregressive way, which represents the approximation to the conditional probability distribution.

In practice, MLP is introduced to perform an affine transformation on the decoder output to constrain the range of parameter values.
For example, for Gaussian likelihood, the parameters can be expressed as $\Phi_t=(\mu_t,\sigma_t)$, where the $\mu_t$ and $\sigma_t$ represent the mean and standard deviation of the likelihood function, respectively.
Then the likelihood and the parameters can be shown in Equation \ref{Eq:gauss} and \ref{Eq:Affine}, respectively.
\begin{equation}
\begin{split}\label{Eq:gauss}
    l(Y_t|\mu_t,\sigma_t)=\frac{exp(-(Y_t-\mu_t)^2/2\sigma_t^2)}{\sqrt{2\pi\sigma_t^2}},
    % \mu_t = W_\mu^T
\end{split}
\end{equation}
\begin{equation}
\begin{split}\label{Eq:Affine}
    \mu_t &= W_\mu^Tf_t+b_\mu,\\
    \sigma_t &= softplus(w_\sigma^Tf_t+b_\sigma),
\end{split}
\end{equation}
where $f_t$ is the output of Transformer at moment $t$.
The softplus function is chosen to ensure that the model generates positive standard deviations.

In the training phase, the historical and predicted sequences are simultaneously input to the model,
where the parallel forecasting is achieved through a masking mechanism.
Owing to the lack of ground truth during inference, 
the forecasted values need to be fed back to the Transformer to achieve sequence forecasting.
We sample $\hat{Y}_{t}$ as the ground truth according to $\hat{Y}_{t}\sim l(Y_t|\mu_t,\sigma_t)$
and feed the samples to the model.

\subsection{Conditional Generative Forecasts}
% In order to achieve hierarchical forecasting and pattern separation for interpretable forecasts, 
% we introduce a conditional generation model to model the primary forecasting result of the Transformer.
In order to mitigate the impact of exposure bias and provide the separation of intricate patterns, 
we introduce a conditional generative model to tackle the primary forecasting result of the Transformer, 
which provides hierarchical forecasting and interpretable forecasting.

% % The hierarchical forecasting represents that the conditional generative model performs non-autoregressive forecasting of 
% % the Transformer prediction results, which brings the forecasted distribution closer to the true distribution of the model.
% The hierarchical forecasting represents that the conditional generative model performs non-autoregressive forecasting of 
% the Transformer prediction results, which brings the forecasted distribution closer to the true distribution.
% Pattern separation is implemented by the sequence decomposition module of the probabilistic decoder, 
% which reconstructs trends and seasonal terms from latent space.
% In addition, the primary forecasting results provided by the Transformer 
% enable the interaction of historical sequences and future information, 
% enhancing the information extraction capability of the decomposition module.

Hierarchical forecasting represents that the conditional generative model performs non-autoregressive forecasting of
the Transformer prediction results, which brings the forecasted distribution closer to the true distribution of the model. 
Pattern separation is implemented by the sequence decomposition module of the probabilistic decoder, which reconstructs trend and seasonality terms from latent space. 
In addition, the primary forecasting results provided by the Transformer enable the interaction of historical sequences and future information, enhancing the information extraction capability of the decomposition module.

Specifically, we are interested in the probabilistic model parameterized by $\theta$ of the form:
\begin{equation}\label{Prob-form}
\begin{split}
% &p_\theta(Y_{t_0+1:t_0+\tau}|Y_{1:t_0},\mu_{t_0+1:t_0+\tau})\\
% =&\int_z p_\theta(Y_{t_0+1:t_0+\tau}|z)p_\theta(z|Y_{1:t_0},\mu_{t_0+1:t_0+\tau})dz
&p_\theta(Y_{t_0+1:t_0+\tau}|Y_{1:t_0})\\
=&\int_z p_\theta(Y_{t_0+1:t_0+\tau}|z)p_\theta(z|Y_{1:t_0})dz,
\end{split}
\end{equation}
% where $\mu_{t_0+1:t_0+\tau}$ represents the likelihood parameters from moment $t_0+1$ to $t_0+\tau$,
where $Y_{t_0+1:t_0+\tau}$ represents the target series from moment $t_0+1$ to $t_0+\tau$,
$z$ represents the latent variables, 
and $p_\theta(z|Y_{1:t_0})$ represents the prior distribution.

The marginal likelihood and posterior density are intractable that require approximate posterior inference necessarily.
We follow the framework of stochastic variational inference \cite{kingma2013auto,sohn2015learning} 
and suppose that the variational posterior is parameterized by $\phi$.
Then the evidence lower bound (ELBO) of the model can be written as follows:
\begin{equation}\label{ELBO}
\begin{split} 
&\textnormal{log}~p_\theta(Y_{t_0+1:t_0+\tau}|Y_{1:t_0})\\
\geq&-D_{KL}\big(q_\phi(z|Y_{1:t_0},Y_{t_0+1:t_0+\tau})||p_\theta(z|Y_{1:t_0})\big) + \\
&\mathbb{E}_{q_\phi(z|Y_{1:t_0},Y_{t_0+1:t_0+\tau})} \big[\textnormal{log}~p_\theta(Y_{t_0+1:t_0+\tau}|z,Y_{1:t_0}) \big],
% \int_z p_\theta(\mu_{t_0+1:t_0+\tau}|z)p_\theta(z|Y_{1:t_0})dz,
\end{split}
\end{equation}
where $q_\phi$ is an approximation of the true posterior and $D_{KL}$ represents the Kullback-Leibler (KL) divergence.
The derivation for the ELBO can be found in Appendix.
Besides, we choose a factor $\beta$ for the KL term to balance capacity of latent variables and independence constraints with reconstruction accuracy, following relevant work \cite{higgins2016beta}.

However, the target series $Y_{t_0+1:t_0+\tau}$ is unknown under the forecasting scenario, which hinders the optimization of the above problem.
Therefore, the generative model in this work actually optimizes the likelihood function output by the Transformer
and reconstructs the probability distribution that is close enough to the true distribution.
Then the ELBO can be denoted as follows:
\begin{equation}\label{ELBO2}
    \begin{split} 
    ELBO=&-D_{KL}\big(q_\phi(z|Y_{1:t_0},\mu_{t_0+1:t_0+\tau})||p_\theta(z|Y_{1:t_0})\big) \\
    &+\mathbb{E}_{q_\phi(z|Y_{1:t_0},Y_{t_0+1:t_0+\tau})} \big[\textnormal{log}~p_\theta(\mu_{t_0+1:t_0+\tau}|z,Y_{1:t_0}) \big],
    % \int_z p_\theta(\mu_{t_0+1:t_0+\tau}|z)p_\theta(z|Y_{1:t_0})dz,
    \end{split}
\end{equation}
where the $\mu_{t_0+1:t_0+\tau}$ represents the parameters of the likelihood function from moment $t_0+1$ to $t_0+\tau$.

\subsubsection{Inference Model}
% For the generative forecast model, likelihood parameters are first mapped by the inference model into latent variable space, 
% where the input sequence can be represented by latent variables.
For the generative forecast model, the likelihood parameters are first mapped by the inference model into the latent variable
space, where the input sequence can be represented by the latent variables.

The inference model (also known as recognition model) $q_\phi(z|Y_{1:t_0},\mu_{t_0+1:t_0+\tau})$ is a conditional Bayesian network,
which is adopted to approximate the intractable posterior distribution $p_\theta(z|Y_{1:t_0})$.
Besides, the KL divergence is applied to assess the similarity between the inference model and the true posterior.
% In this work, the likelihood parameters of the autoregressive forecasting are taken as the modeling target, 
% and the output of the generative model is the reconstructed likelihood parameters.
As shown in the right of Figure \ref{Fig:PDT}, the inferential model can be implemented by the probabilistic encoder.
We utilize MLP to achieve the probabilistic encoder, 
where the input of the probabilistic encoder consists of the historical series $Y_{1:t_0}$ and 
the parameters of the likelihood function $\mu_{t_0+1:t_0+\tau}$.
% the target series $Y_{t_0+1:t_0+\tau}$.
% However, the target series is unknown under the forecasting scenario, 

The probabilistic encoder outputs the parameters of a Gaussian distribution and samples the latent variable z by a reparameterization trick \cite{kingma2015variational}.
The reparameterization trick converts the representation of the random variable $z$ into a deterministic part and a stochastic part, shown as follows:
\begin{equation}\label{reparameterization}
\begin{split} 
z= \mu + \sigma\odot\varepsilon,\\
\varepsilon\sim\mathcal{N}(0,1),
\end{split}
\end{equation}
where $\mu$ and $\sigma$ are parameters of Gaussian distribution generated by the probabilistic encoder, 
$\varepsilon$ is random variable sampled by standard Gaussian distribution.

\subsubsection{Generative Model}
The generative model refers to the generation of forecasting values according to observed variables and latent variables, which can be expressed as $p_\theta(\mu_{t_0+1:t_0+\tau}|z,Y_{1:t_0})$.

In this work, the generator is employed to reconstruct the likelihood parameters, 
and we want the reconstructed likelihood function to be close enough to the true conditional distribution.
Different from the reconstructed input of the original VAE, the model in this paper aims at approximating the ground truth.
Therefore, we choose the negative log-likelihood function as the reconstruction loss in the generative model.

In addition, another task of the generative model is pattern separation for interpretable prediction.
In the context of forecasts,  the probabilistic encoder maps the primary forecasts result to the latent space, 
which can compensate for the lack of future information and improve the performance of the model to decompose the target series.
The probability decoder reconstructs typical features of the sequence, such as seasonality and trend terms, 
% from probability distributions in the latent space to achieve patterns separation and provide interpretable forecasts.
from Gaussian distributions in the latent space to achieve pattern separation and provide interpretable forecasts,
where the historical sequence is fed into the decoder as conditional information.
As shown in the right of Figure \ref{Fig:PDT}.

Let $\hat{\mu}_t$ denotes the reconstructed likelihood parameters corresponding to the target time series at time $t$, 
where $Y_t \sim \mathcal{N}(\hat{\mu}_t,\sigma_t^2)$.
We introduce the probabilistic decoder to decompose the above distribution into two Gaussian distributions with the following characteristics:
\begin{equation}\label{De}
    \begin{split} 
&Y^{trend}_t \sim \mathcal{N}(\mu^{trend}_t,\sigma^2_t/2),\\
&Y^{seasonal}_t \sim \mathcal{N}(\mu^{seasonal}_t,\sigma^2_t/2),\\
&\hat{\mu}_t = \mu^{trend}_t+\mu^{seasonal}_t,
    \end{split}
\end{equation}
where $\mu^{trend}_t$ and $\mu^{seasonal}_t$ represent likelihood parameters of the trend term and seasonality term at time $t$, respectively.

Specifically, the probability decoder takes the latent variable as input and obtains the decomposition features 
through the trend feature extraction and seasonality feature extraction modules.
The trend term can be extracted explicitly by convolution operation that can simulate the moving average to smooth out periodic
fluctuations and highlight the trends.
The prediction results are obtained by linear decoding of trend term and seasonality term, where the seasonality term can be extracted implicitly by MLP.
Let AvgPooling represents the average pooling operation for moving average and LinearDecoder denotes a linear decoder, 
then the process is shown as follows:
\begin{equation}\label{De}
    \begin{split} 
    \mu^{trend}_{t_0+1:t_0+\tau} =& AvgPooling\big(MLP(z)\big),\\
    \mu^{seasonal}_{t_0+1:t_0+\tau} =& MLP \big(z\big),\\
    % \hat{\mu}_{t_0+1:t_0+\tau}  =& \mu^{trend}_{t_0+1:t_0+\tau} + \mu^{seasonal}_{t_0+1:t_0+\tau},
    \hat{\mu}_{t_0+1:t_0+\tau}  =& LinearDecoder \big(\mu^{trend}_{t_0+1:t_0+\tau} , \mu^{seasonal}_{t_0+1:t_0+\tau} \big).
    \end{split}
\end{equation}

Then, given likelihood parameters $\hat{\mu}_t$ and $\sigma_t$, the forecasting result can be obtained by sampling the Gaussian distribution 
$\hat{Y}_t \sim \mathcal{N}(\hat{\mu}_t,\sigma_t^2)$.

\subsection{Joint Learning}
We define several loss terms to train the proposed model jointly, where the Transformer contains the negative log-likelihood (NLL) loss and the conditional generative model contains the KL divergence and reconstruction loss.
\subsubsection{Negative Log-Likelihood loss}
% ($\mathcal{L}_{NLL}$)
The Transformer predicts the conditional distributions by maximum likelihood estimation.
The negative log-likelihood is chosen as the loss function, where the minimization loss function is equivalent to the maximum likelihood estimation.
For Gaussian likelihood, the loss of NLL is shown as follows:
\begin{equation}\label{NLL}
\begin{split} 
\mathcal{L}_{NLL} =& \sum_t l(Y_t|\mu_t,\sigma_t)
= \sum_t \frac{(Y_t-\mu_t)^2}{2\sigma_t^2}+log~\sigma_t +Const,
\end{split}
\end{equation}
where the detail can be found in \cite{salinas2020deepar}.
\subsubsection{KL loss}
% ($\mathcal{L}_{KL}$)
As in VAE, we regularize the latent space by encouraging it to be similar to a standard Gaussian
distribution with $\mu$ the null vector and $\sigma$ the identity matrix. 
We minimize the KL divergence between the encoder distribution and the true posterior.
\begin{equation}\label{KL}
\begin{split} 
% \mathcal{L}_{KL} =& -D_{KL}\big(q_\phi(z|Y_{1:t_0},\mu_{t_0+1:t_0+\tau})||p_\theta(z|Y_{1:t_0})\big)\\
% =& \frac{1}{2}\sum_{t=t_0+1}^{t_0+\tau}\big(1+log~\sigma_t^2-\mu_t^2-\sigma_t^2\big).
\mathcal{L}_{KL} =& -D_{KL}\big(q_\phi(z|Y_{1:t_0},\mu_{t_0+1:t_0+\tau})||p_\theta(z|Y_{1:t_0})\big)\\
=& \frac{1}{2}\sum_{j=1}^{N}\big(1+log~\sigma_j^2-\mu_j^2-\sigma_j^2\big),
\end{split}
\end{equation}
where $N$ is the number of the latent variables, $\mu_j$ and $\sigma_j$ are element of $\mu$ and $\sigma$, respectively.

\subsubsection{Reconstruction loss}
One of the goals of the generative model is to maximize the expectation term in evidence lower bound, i.e., 
$maximum \  \mathbb{E}_{q_\phi(z|Y_{1:t_0},\mu_{t_0+1:t_0+\tau})} \big[\textnormal{log}~p_\theta(\mu_{t_0+1:t_0+\tau}|z,Y_{1:t_0}) \big]$.
The maximum problem is equivalent to minimizing the reconstruction loss.
For the probabilistic decoder, the reconstruction loss represents the error of the generative model output with respect to the likelihood function, which can be denoted as $\mathcal{L}_{R}(\hat{\mu_t},\mu_t)$.
Considering that the likelihood function obtained from the Transformer is used to approximate the true distribution, we choose to directly measure the loss between the likelihood of the reconstruction and the true distribution, which can be denoted as $\mathcal{L}_{R}(\hat{\mu_t},Y_t)$.
In experiments, $\mathcal{L}_{R}(\hat{\mu_t},Y_t)$ can be chosen as the NLL, the above scheme was observed to achieve better forecasting performance.
% The reconstruction loss is the expectation term in evidence lower bound, i.e., 
% $\mathcal{L}_{R}=\mathbb{E}_{q_\phi(z|Y_{1:t_0},\mu_{t_0+1:t_0+\tau})} \big[\textnormal{log}~p_\theta(\mu_{t_0+1:t_0+\tau}|z,Y_{1:t_0}) \big]$.
% For the probabilistic decoder, the $\mathcal{L}_{R}$ can be chosen as MSE or NLL.

The total loss of the PDTrans is defined as the weighted summation of different terms: $\mathcal{L}=\gamma \mathcal{L}_{NLL}+\beta \mathcal{L}_{KL}+\mathcal{L}_{R}$, where $\beta$ and $\gamma$ are trade-off coefficients.

\section{Experiments}

% \subsection{Datasets And Evaluation Metrics}
\subsection{Datasets}

In this work, five public datasets are used for model performance evaluation, i.e., Electricity, Traffic, Solar, Exchange, and M4-Hourly.
The general statistics are listed in Table \ref{Tab:Datasets}.

\begin{table}[htb]
    \renewcommand\arraystretch{1.2}  
    \centering
    % \scriptsize
    % \normalsize
    \small
    \caption{General information about datasets, where the 3rd column represents the sequence length of each variable. }
    \begin{tabularx}\linewidth{XXXX}
        \toprule 
        Datasets & Resolution &  Length & Variables\\
        \midrule
        Electricity & 1 hour&32,304&370\\
        Traffic &1 hour&4,049&963\\
        Solar &1 hour&4,832&137\\
        Exchange & 1 day& 7,588&8\\
        M4-Hourly &1 hour&748/1,008&414\\
        \bottomrule   
    \end{tabularx}
    \label{Tab:Datasets}
    % }
\end{table}

\begin{table}[htb]
    \centering
    \footnotesize
    \caption{Hyperparameters for PDTrans, where $N$ is number of layers, $h$ is number of head, $d_{model}$ is dimensionality of model, $d_{ff}$ is dimensionality of inner-layer, $\lambda$ and $\beta$ represent trade-off coeﬀicients, $k$ is kernel size of 1-D convolutions.}
        \begin{tabularx}\linewidth{p{1.5cm}XXXXXXX}
            \toprule 
            &$N$ &$h$ &$d_{model}$ &$d_{ff}$ &$\lambda$ & $\beta$ &$k$\\
            \midrule
            Electricity &3&8&160&2048&1&1&5\\
            Traffic &3&8&160&2048&1&1&3\\
            Solar &3&8&16&640&1&1&3\\
            Exchange &3&4&160&640&1&1&5\\
            M4-Hourly &3&8&160&2048&1&1&3\\
            \bottomrule   
        \end{tabularx}
        \label{Tab:Para}
\end{table}

The Electricity dataset consists of the electricity consumption of 370 users at a 15-minute resolution from 01/01/2011 to 07/09/2014, where in this work all series are aggregated into hourly intervals.
The Traffic dataset is composed of the road occupancy rate, between 0 and 1, where the dataset contains 963 related variables with hourly resolution from 02/01/2008 to 30/03/2009.
The Solar dataset consists of hourly interval solar power production data collected from 137 photovoltaic plants in Alabama from January to August 2006.
The Exchange dataset contains daily exchange rate records from January 1990 to December 2016 for 8 countries.
The M4-Hourly dataset includes 414 time series of hourly intervals from M4 competition \cite{makridakis}, where the training and test set have been provided.

For covariates, we consider both static features and dynamic features,
where the static features refer to the serial ID 
% The static features refer to the serial ID, where
% time-independent features, where 
and the dynamic features refer to the time-related features such as time period and relative positions.
% The relative positions can be described as the distance to the first value of the dataset following \cite{li2019enhancing,salinas2020deepar}.
The time period contains the encoding of hour and day for hourly interval datasets, and the encoding of day and month for daily interval datasets.
The relative positions can be described as the distance to the first value of the series for all datasets.

\begin{table*}[htb]
\centering
\footnotesize
\caption{$\rho_{0.5}/\rho_{0.9}$ metrics for the short-term (1d ahead) and long-term (7d ahead) forecasting scenarios on Electricity and Traffic datasets for autoregressive probabilistic forecasting models. $\Diamond$ denotes results from \cite{lin2021ssdnet}.}
% We only report $\rho_{0.5}$ for points predictions models.
    \begin{tabularx}\linewidth{XXXXXX}
        \toprule 
         &DeepAR & DeepSSM  &ConvTrans &  SSDNet& \textbf{PDTrans(our)}\\
        \midrule
         Elect$_{1d}$& 0.075$^\Diamond$ / 0.040$^\Diamond$& 0.083$^\Diamond$ / 0.056$^\Diamond$  &0.059$^\Diamond$ / 0.034$^\Diamond$&  0.068$^\Diamond$ / 0.033$^\Diamond$ & \textbf{0.058} / \textbf{0.030}\\
         Elect$_{7d}$& 0.082$^\Diamond$ / 0.053$^\Diamond$& 0.085$^\Diamond$ / 0.052$^\Diamond$& 0.070$^\Diamond$/ 0.044$^\Diamond$ & 0.079 / 0.042 & \textbf{0.068} / \textbf{0.038}\\
         Traffic$_{1d}$& 0.161$^\Diamond$ / 0.099$^\Diamond$& 0.167$^\Diamond$ / 0.113$^\Diamond$& 0.122$^\Diamond$ / \textbf{0.081}$^\Diamond$ &0.153 / 0.101 & \textbf{0.113} / 0.088\\
         Traffic$_{7d}$&  0.179$^\Diamond$ / 0.105$^\Diamond$ & 0.168$^\Diamond$ / 0.114$^\Diamond$ & 0.139$^\Diamond$ / 0.094$^\Diamond$& 0.161 / 0.109 & \textbf{0.126} / \textbf{0.093} \\
        \bottomrule   
    \end{tabularx}
    \label{Tab:Exp1}
\end{table*}

\begin{table*}[htb]
\centering
\footnotesize
\caption{$\rho_{0.5}/\rho_{0.9}$ metrics of various methods on five real datasets. $\Diamond$ indicates results reported by \cite{lin2021ssdnet}.}
    \begin{tabularx}\linewidth{XXXXXXXXX}
        \toprule 
        &Electricity&Traffic & Solar&Exchange&M4-Hourly \\
        \midrule
        Prophet & 0.112 / 0.055&  0.183 / 0.137   &0.268 / 0.169   &  0.017 / 0.013   &  0.102 / 0.038  \\
        DeepAR &  0.075$^\Diamond$ / 0.040$^\Diamond$& 0.161 / 0.099 & 0.222$^\Diamond$ / 0.093$^\Diamond$ & 0.014$^\Diamond$ / 0.009$^\Diamond$ & 0.090 / 0.030\\
        DeepSSM & 0.083$^\Diamond$ / 0.056$^\Diamond$& 0.167 / 0.133 & 0.223$^\Diamond$ / 0.181$^\Diamond$ & 0.014$^\Diamond$ / 0.012$^\Diamond$& 0.044 / 0.026\\
        ConvTrans & 0.059$^\Diamond$ / 0.034$^\Diamond$ & 0.122$^\Diamond$ / \textbf{0.081}$^\Diamond$ & 0.210$^\Diamond$ / 0.082$^\Diamond$ & 0.017 / 0.008 & 0.067 / 0.025\\
        N-Beats& 0.061$^\Diamond$ / - & 0.114 / - & 0.212 / - & 0.018 / - & \textbf{0.025} / -\\
        % N-Beats-I& 0.061 / - & 0.114 / - & 0.212 / - & 0.018 / - & 0.025 / -\\
        Informer &  0.068$^\Diamond$ / -& 0.122 / - & 0.215$^\Diamond$ / - & 0.014$^\Diamond$ / -&- / -\\
        Autoformer& 0.083 / - & 0.120 / - & 0.211 / - & 0.013 / - &- / -\\
        SSDNet& 0.068$^\Diamond$ / 0.033$^\Diamond$ & 0.166 / 0.106  & 0.209$^\Diamond$ / 0.074$^\Diamond$ & 0.013$^\Diamond$ / \textbf{0.006}$^\Diamond$ &0.038 / \textbf{0.023}\\
        \textbf{PDTrans(our)}& \textbf{0.058} / \textbf{0.030} & \textbf{0.113} / 0.088 & \textbf{0.205} / \textbf{0.073} & \textbf{0.011} / \textbf{0.006} &0.033 / \textbf{0.023}\\
        \bottomrule   
    \end{tabularx}
    \label{Tab:Exp2}
\end{table*}

\begin{table*}[htb]
\centering
\footnotesize
\caption{Ablation study. The performance of $\rho_{0.5}/\rho_{0.9}$ loss. w/ denotes with probabilistic decomposition module and w/o denotes without probabilistic decomposition module.}
    \begin{tabularx}\linewidth{XXXXXXXXX}
        \toprule 
        &Electricity&Traffic & Solar&Exchange&M4-Hourly \\
        \midrule
        PDTrans(w/o) & 0.064 / 0.036 & 0.134 / 0.090 & 0.214 / 0.086 & 0.018 / 0.008 & 0.046 / 0.028\\
        % VAE &  &  &  &  &\\
        % PDT(w/o) &  &  &  &  &\\
        PDTrans(w/)& \textbf{0.058} / \textbf{0.030} & \textbf{0.113} / \textbf{0.088} & \textbf{0.205} / \textbf{0.073} & \textbf{0.011} / \textbf{0.006} &\textbf{0.033} / \textbf{0.023}\\
        \bottomrule   
    \end{tabularx}
    \label{Tab:Abla}
\end{table*}

\begin{table}[htb]
    \centering
    \footnotesize
    \caption{Ablation study on Electricity dataset. The effect of the trade-off coefficients $\beta$ and $\gamma$. }
        \begin{tabularx}\linewidth{XXX}
            \toprule 
            % &$\gamma=0.5,\beta=0.5$&$\gamma=1,\beta=1$&$\gamma=5,\beta=5$ \\
            &$\rho_{0.5}$ & $\rho_{0.9}$ \\
            \midrule
            $\gamma=0.5,\beta=0.5$ & 0.058&0.032\\
            % $\gamma=0.5,\beta=1$ & \\
            $\gamma=1,\beta=1$ & 0.058&0.030\\
            % $\gamma=1,\beta=5$ & \\
            $\gamma=5,\beta=5$ & 0.059&0.031\\
            $\gamma=10,\beta=10$ & 0.064&0.035\\
            % $\rho_{0.5}/\rho_{0.9}$ & & & \\
            % DeepAR & 0.075 / 0.040  & 0.161 / 0.099\\
            % PD-DeepAR & \textbf{0.072} / \textbf{0.036} & \textbf{0.147} / \textbf{0.096}   \\
            \bottomrule   
        \end{tabularx}
        \label{Tab:Abl-coff}
\end{table}

\begin{table}[htb]

\centering
\footnotesize
\caption{Comparing the model of DeepAR and the model of DeepAR equipped with probabilistic decomposition. }
    \begin{tabularx}\linewidth{XXX}
        \toprule 
        &Electricity&Traffic \\
        \midrule
        DeepAR & 0.075 / 0.040  & 0.161 / 0.099\\
        PD-DeepAR & \textbf{0.072} / \textbf{0.036} & \textbf{0.147} / \textbf{0.096}   \\
        \bottomrule   
    \end{tabularx}
    \label{Tab:Exp3}
\end{table}

\subsection{Evaluation Metrics And Experimental Setup}
Following \cite{salinas2020deepar,li2019enhancing}, we evaluate the model using $\rho$-quantile loss with $\rho\in(0,1)$, which is shown as follows:
\begin{equation}
\begin{split}
    Q_\rho(Y,\hat{Y})&=\frac{2\sum_tP_\rho(y_{t},\hat{y}_{t})}{\sum_t|y_{t}|},
    % P_\rho(y_{t},\hat{y}_{t})&=(y_t-\hat{y}_t)\big(\rho I_{\hat{y}_t>y_t}-(1-\rho)I_{\hat{y}_t\leq y_t}\big),
\end{split}
\end{equation}
\begin{equation}
\begin{split}
    % Q_\rho(Y,\hat{Y})&=\frac{2\sum_tP_\rho(y_{t},\hat{y}_{t})}{\sum_t|y_{t}|},\\
    P_\rho(y_{t},\hat{y}_{t})&=(y_t-\hat{y}_t)\big(\rho I_{\hat{y}_t>y_t}-(1-\rho)I_{\hat{y}_t\leq y_t}\big),
\end{split}
\end{equation}
where $y_t$ is the ground truth, $\hat{y}_t$ represents the predicted distribution with $\rho$-quantile, and $I$ is a boolean function.
We report $\rho_{0.5}$ and $\rho_{0.9}$ metrics to be consistent with the literature \cite{salinas2020deepar,li2019enhancing}.

We utilize Vaswani Transformer \cite{vaswani2017attention} for primary forecasts, where the number of layers of both encoder and decoder is set to 3, and the head of the attention mechanism is set to 4 and 8 for different datasets.
The dimensionality of the inner layer $d_{ff}$ is set to 2048 for Electricity, Traffic, and M4-Hourly datasets, and set to 640 for Solar and Exchange datasets.
The dimensionality of the model $d_{model}$ is set to 16 for the Solar dataset and set to 160 for the others.
For the primary forecasts model, we utilize learnable position and covariates embedding.
The embedding dimensionality is set to 20 and 30.
For the generative model, both of the trade-off coefficients $\beta$ and $\gamma$ are set to 1 for all datasets.
The dimensionality of latent space is set to 20 for all datasets.
The kernel size of the 1-D convolutions in the trend extraction module is set to 3 and 5 for different datasets.
The main hyperparameters used in this work are listed in Table \ref{Tab:Para}.

Furthermore, the split of the datasets is followed by the literature \cite{salinas2020deepar,li2019enhancing,lin2021ssdnet}.
We utilize one week of data from 9/1/2014 on Electricity and 6/16/2008 on Traffic as test sets.
For the Solar dataset, the last 7 days in August are used as test set.
In addition, we leave one week of data for the validation set and the rest of the data as the training set for the above datasets.
% For the Exchange dataset, the working day from 2/2/2015 is used as test set, the 480 working
% days before 2/2/2015 are set to validation set, and the remaining data is used as training set. 
For the Exchange dataset, only the weekday data is considered, where 90\% of the data before February 2, 2015 is used as the training set and 10\% of the data is used as the validation set. 
The rest of the data after February 2, 2015 is used as the test set.
% The split of the M4-Hourly dataset is already provided.
The M4-Hourly dataset has been provided with a division of the training, validation, and test sets.

For all datasets, the Adam \cite{kingma2014adam} optimizer is employed to optimize the model with a learning rate of 0.001, where the learning rate is set to decrease by 20\% every two epochs.
Besides, we set the maximum number of epochs to 100 for all experiments

In this work, all experiments are carried out with PyTorch 1.8 on NVIDIA RTX 3090 GPU in the Ubuntu22.04 environment.
Codes will be available after acceptance at https://github.com/JL-tong/PDTrans.git.

\begin{figure}[!htb]
    \centering
    \includegraphics[width=0.5\textwidth]{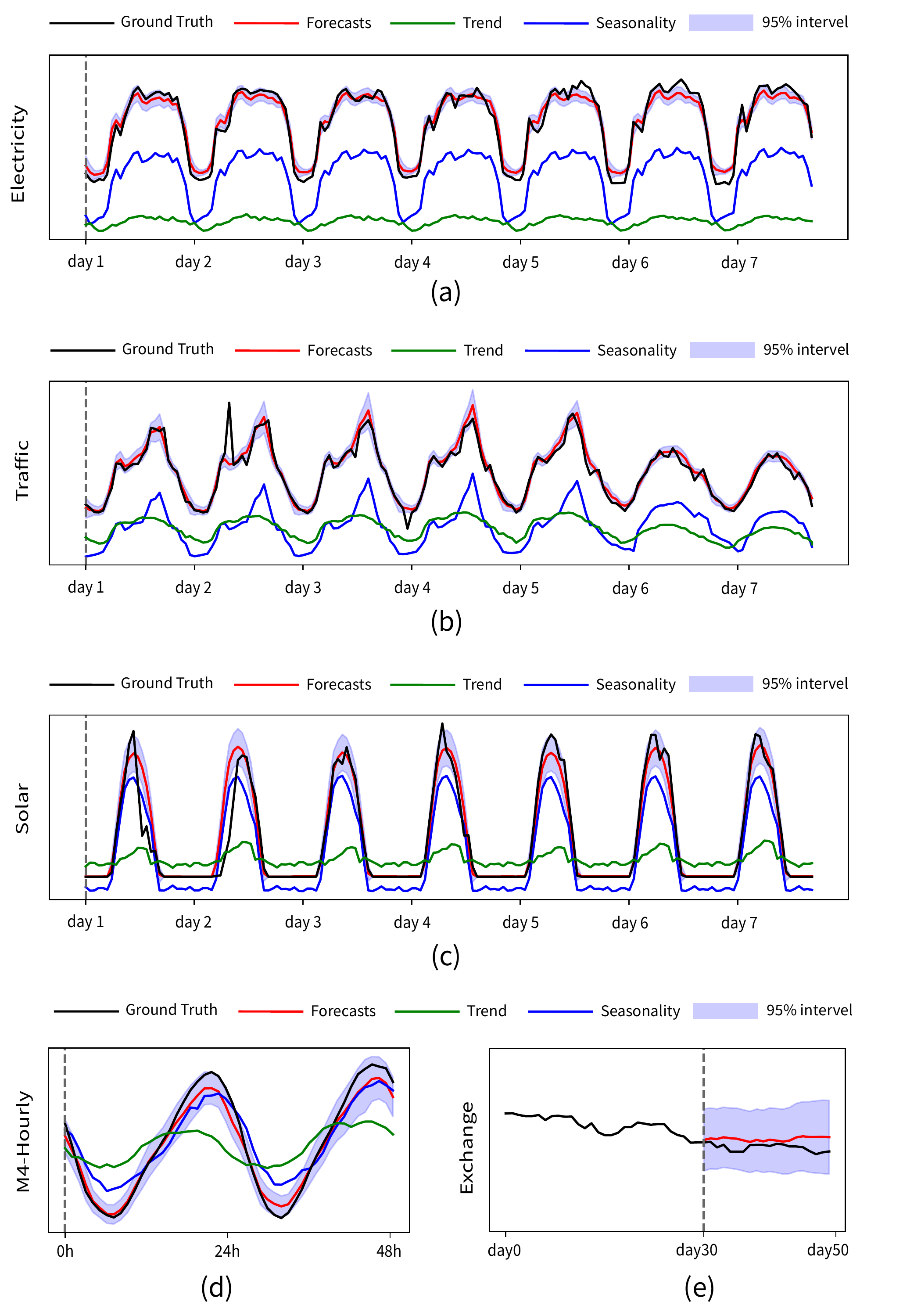}\\
    \caption{The forecasting results. (a) Result on Electricity. (b) Result on Traffic.(c) Result on Solar. (d) Result on M4-Hourly. (e) Result on Exchange. }
    \label{Fig:Result}
\end{figure}

\subsection{Accuracy Analysis}

We first evaluate the performance of long-term and short-term forecasting on Electricity and Traffic datasets with the probabilistic decomposition Transformer model.
Both the Electricity and Traffic datasets are divided into 7 days of data for testing. 
For short-term forecasts, we follow \cite{li2019enhancing} to assess the performance by rolling forecasts for 7 days, with each forecast being 24 hours long.
% evaluate rolling-day forecasts for 7 days following \cite{li2019enhancing}. 
For the long-term forecasting, the length of prediction range is specified as 7 days directly, with 14 days (i.e., 336 observations per time series) for conditioning range.
The comparison methods are autoregressive probabilistic models include DeepAR\cite{salinas2020deepar}, DeepSSM\cite{rangapuram2018deep}, ConvTrans\cite{li2019enhancing},and SSDNet\cite{lin2021ssdnet}.
The result is shown in Table \ref{Tab:Exp1}.

Our model achieves the best performance for both long-term and short-term forecasting on Electricity dataset.
Besides, the PDTrans get better results in long-term forecasting both for $\rho_{0.5}$ and $\rho_{0.9}$ on Traffic dataset.
For short-term forecasting on Traffic dataset, the proposed model achieved the best $\rho_{0.5}$ and a competitive $\rho_{0.9}$.
The results indicate that the hierarchical forecasting mechanism can effectively keep the long-term dependency of the model and improve forecasting performance.

To more comprehensively assess the effectiveness of the proposed model, 
the PDTrans is compared with the mainstream methods on five real-world datasets.
% we compare our model on five real-world datasets with the mainstream methods.
For the hourly interval dataset, the model input length is set to 7 days and the prediction length is specified as 24 hours.
For the day-interval dataset, the model input is set to 30 and the prediction length is selected as 20, following \cite{lin2021ssdnet}.
The comparison methods include Prophet \cite{taylor2018forecasting}, DeepAR\cite{salinas2020deepar}, DeepSSM\cite{rangapuram2018deep}, ConvTrans\cite{li2019enhancing}, N-Beats\cite{oreshkin2019n}, Informer\cite{zhou2021informer}, Autoformer\cite{xu2021autoformer}, and SSDNet\cite{lin2021ssdnet}.
The $\rho_{0.5}$ and $\rho_{0.9}$ metrics are reported for probabilistic forecasting models.
Besides, we only report $\rho_{0.5}$ metrics for the non-probabilistic forecasting models.

For Electricity, Solar, and Exchange datasets, the proposed PDTrans has the best $\rho_{0.5}$ and $\rho_{0.9}$ among the models involved in the comparison.
% among the probabilistic forecasting models and the interpretable models (such as DeepSSM, N-Beats).
Besides, our method achieves the best performance for $\rho_{0.5}$ metrics on Traffic dataset, and a very competitive performance for the $\rho_{0.9}$ metrics.
The PDTrans performs better than autoregressive probabilistic forecasting models, such as DeepAR and ConvTrans, which indicates that hierarchical probabilistic forecasting can improve the performance of autoregressive probabilistic forecasting.
In addition, our model performs well on datasets with significant periodicities such as the Electricity and Solar datasets, indicating that the model has an excellent ability for mining periodic features.

\subsection{Interpretability Analysis}

The model in this paper extracts the short-term trend instead of the long-term trend since the long-term trend requires obtaining a long series, which is difficult to operate in practice.

Figure \ref{Fig:Result} presents the results of a rolling 7-day forecasting on the Electricity, Traffic, and Solar datasets, where the black line represents the ground truth, the red line denotes the forecasting result, the green line is the trend term of forecasting result, and the blue line is the seasonality term of forecasting result.
The purple shaded area of the figure represents the 95\% confidence interval.
For the M4-Hourly dataset, only the 48-hour ahead forecasting results are presented in the figure.
Besides, the decomposition results of the Exchange dataset are not shown, due to the lack of regular fluctuations.

The results in Figure \ref{Fig:Result} are interpretable that the trend curve is moving proposed and reflects the trend of change, 
the seasonality term presents regular fluctuations.
For instance, in Figure \ref{Fig:Result} (b)  from day 5 to day 7, the green line not only presents the daily trend, but also reflects the decreasing trend of the peak.

\subsection{Ablation Study}
We conducted ablation experiments to evaluate the effectiveness of the probabilistic decomposition module part of PDTrans.
Specifically speaking, we compare the PDTrans with the model eliminating the probability decomposition module that is equivalent to the autoregressive probabilistic Transformer.
The short-term forecasting performance of the two models on the five datasets is shown in Table \ref{Tab:Abla}.
For $\rho_{0.5}$ metrics, the PDTrans decreases by 9\%, 15\%, 4\%, 38\%, 28\% for the Electricity, Traffic, Solar, Exchange, and M4-Hourly, respectively.
A similar phenomenon can be observed for the $\rho_{0.9}$ metrics.
The results indicate that the model equipped with the probabilistic decomposition module achieves better performance.

In addition, to analyze the effect of different trade-off coefficients on the model, we designed relevant experiments on the Electricity dataset.
The result can be found in Table \ref{Tab:Abl-coff}.
The experimental results show that the model performance fluctuates slightly when the coefficients fluctuate within a certain range. 
% When both $\beta$ and $\gamma$ are selected as 1, the $\rho_{0.5}$ and $\rho_{0.9}$ evaluation metrics are 0.058 and 0.030, respectively.
% However, the $\rho_{0.5}$ and $\rho_{0.9}$ evaluation metrics are 0.059 and 0.031, respectively, when $\beta=5$ and $\gamma=5$.
For example, when the coefficients $\beta$ and $\gamma$ both change from 1 to 5, the $\rho_{0.5}$ evaluation indicator increases by only 1.69\% from 0.058 to 0.059 and the $\rho_{0.9}$ increases by 3.23\% from 0.030 to 0.031.
% This indicates that the PDTrans model is not sensitive to the trade-off coefficients within a certain range.
The results show that the PDTrans model is not sensitive to the trade-off coefficients within a certain range, demonstrating satisfactory robustness.

\subsection{Further Exploration}

To further explore the benefits of the probabilistic decomposition module to time series forecasting tasks,
the probabilistic decomposition module is employed to the DeepAR \cite{salinas2020deepar}, 
% we attach the probabilistic decomposition module to the DeepAR \cite{salinas2020deepar}, 
which is an autoregressive probabilistic forecasting network based on LSTM.
Similar to PDTrans, the probabilistic decomposition module is employed to model the likelihood that output by LSTM for hierarchical probabilistic and interpretable forecasting.
The forecasting result on Electricity and Traffic datasets can be found in Table \ref{Tab:Exp3}.
The forecasting result shown in Table \ref{Tab:Exp3} indicates that the PD-DeepAR performs significantly better than DeepAR on both datasets.
It indicates that the probability decomposition module can effectively improve the forecasting performance of the model through hierarchical forecasting.

\section{Conclusion}
In this work, we propose probabilistic decomposition Transformer (PDTrans) model for hierarchical and interpretable probabilistic forecasting of intricate time series data, where the PDTrans consists of Transformer and conditional generative model.
The Transformer is employed to extract temporal patterns and implement primary autoregressive probabilistic forecasting.
% The conditional generative model achieves hierarchical probabilistic forecasting through variational inference, which can effectively reduce the impact of exposure bias in the autoregressive forecasting process.
The conditional generative model achieves hierarchical probabilistic forecasting through variational inference, which can effectively reduce the impact of exposure bias in the autoregressive forecasting process.
In addition, the conditional generative model generates trends and seasonality features by probabilistic decoders to achieve separation of intricate patterns and interpretable forecasts.
A series of ablation experiments are designed to demonstrate the effectiveness and robustness of the probabilistic decomposition block.
Moreover, the performance of the PDTrans is evaluated on five time series datasets, showing that it compares favorably with state of the art in terms of accuracy.
The results indicate that the PDTrans is a reliable alternative to time series forecasting.

\section*{Acknowledgements} 
This work was supported in part by the Zhishan Young Scholar Program of Southeast University; in part by the Fundamental Research Funds for the Central Universities under Grant 2242021R41118. Besides, we thank the Big Data Computing Center of Southeast University for providing the facility support on the numerical calculations. 
% \section*{Appendix}\appendix
\appendix

\section{Evidence Lower Bound of Log-Likelihood}
% \subsubsection*{Variational Lower Bound of Log-Likelihood}
The derivation for the evidence lower bound of log-likelihood is denoted as follows:
\begin{equation}\label{ELBO-proof}
    \begin{split} 
    &\textnormal{log}~p_\theta(\mu_{t_0+1:t_0+\tau}|Y_{1:t_0})\\
    =&D_{KL}\big(q_\phi(z|Y_{1:t_0},\mu_{t_0+1:t_0+\tau})||p_\theta(z|Y_{1:t_0},\mu_{t_0+1:t_0+\tau})\big) \\
    &+ \mathbb{E}_{q_\phi(z|Y_{1:t_0},\mu_{t_0+1:t_0+\tau})} \big[-\textnormal{log}~q_\phi(z|Y_{1:t_0},\mu_{t_0+1:t_0+\tau}) \big]\\
    &+\mathbb{E}_{q_\phi(z|Y_{1:t_0},\mu_{t_0+1:t_0+\tau})} \big[\textnormal{log}~p_\theta(\mu_{t_0+1:t_0+\tau},z|Y_{1:t_0}) \big]\\
    \geq& \mathbb{E}_{q_\phi(z|Y_{1:t_0},\mu_{t_0+1:t_0+\tau})} \big[-\textnormal{log}~q_\phi(z|Y_{1:t_0},\mu_{t_0+1:t_0+\tau}) \big]\\
    &+\mathbb{E}_{q_\phi(z|Y_{1:t_0},\mu_{t_0+1:t_0+\tau})} \big[\textnormal{log}~p_\theta(\mu_{t_0+1:t_0+\tau},z|Y_{1:t_0}) \big]\\
    =& \mathbb{E}_{q_\phi(z|Y_{1:t_0},\mu_{t_0+1:t_0+\tau})} \big[-\textnormal{log}~q_\phi(z|Y_{1:t_0},\mu_{t_0+1:t_0+\tau}) \big]\\
    &+\mathbb{E}_{q_\phi(z|Y_{1:t_0},\mu_{t_0+1:t_0+\tau})} \big[\textnormal{log}~p_\theta(z|Y_{1:t_0}) \big]\\
    &+\mathbb{E}_{q_\phi(z|Y_{1:t_0},\mu_{t_0+1:t_0+\tau})} \big[\textnormal{log}~p_\theta(\mu_{t_0+1:t_0+\tau}|Y_{1:t_0},z) \big]\\
    =&-D_{KL}\big(q_\phi(z|Y_{1:t_0},\mu_{t_0+1:t_0+\tau})||p_\theta(z|Y_{1:t_0})\big) + \\
    &\mathbb{E}_{q_\phi(z|Y_{1:t_0},\mu_{t_0+1:t_0+\tau})} \big[\textnormal{log}~p_\theta(\mu_{t_0+1:t_0+\tau}|z,Y_{1:t_0}) \big].
    \end{split}
\end{equation}

\balance

\end{document}